\documentclass[letterpaper, 10 pt, conference]{ieeeconf}  

\usepackage{dsfont}
\usepackage{multicol}
\usepackage{booktabs}
\usepackage{graphicx}
\usepackage{xcolor}
\usepackage{amsmath, amsfonts, amssymb}
\usepackage{float}
\usepackage{pgfplots}
\usepgfplotslibrary{groupplots, colormaps, statistics, fillbetween}
\usepackage{multirow}
\usepackage{makecell}
\usepackage{rotating}
\usepackage[linesnumbered,ruled,vlined]{algorithm2e}
\usepackage{arydshln} 
\usetikzlibrary{arrows.meta}
\usepackage{adjustbox}
\usepackage[normalem]{ulem} 
\usepackage[bookmarks=false]{hyperref}
\usepackage{balance}

\def\gZ{{\mathcal{Z}}}
\newtheorem{defn}{Definition}
\newtheorem{thm}[defn]{Theorem}

\pgfplotsset{compat=1.18}

\renewenvironment{thebibliography}[1]{
  \begin{oldthebibliography}{#1}
    \setlength{\itemsep}{0em}
    \setlength{\parskip}{0em}
}
{
  \end{oldthebibliography}
}

\IEEEoverridecommandlockouts                              

\title{\LARGE \bf SLIM-VDB: A Real-Time 3D Probabilistic\\Semantic Mapping Framework}

\author{\authorblockN{Anja Sheppard\authorrefmark{1},
Parker Ewen\authorrefmark{1},
Joey Wilson\authorrefmark{1},
Advaith V. Sethuraman\authorrefmark{1}, \\
Benard Adewole\authorrefmark{1},
Anran Li\authorrefmark{1},
Yuzhen Chen\authorrefmark{1},
Ram Vasudevan\authorrefmark{1} and
Katherine A. Skinner\authorrefmark{1}}
\authorblockA{\authorrefmark{1} Department of Robotics \\
University of Michigan,
Ann Arbor, MI, USA 48104 \\ Email: anjashep@umich.edu}}

\begin{document}

\maketitle
\thispagestyle{empty}
\pagestyle{empty}

\begin{abstract}

This paper introduces SLIM-VDB, a new lightweight semantic mapping system with probabilistic semantic fusion for closed-set or open-set dictionaries.
Advances in data structures from the computer graphics community, such as OpenVDB, have demonstrated significantly improved computational and memory efficiency in volumetric scene representation. 
Although OpenVDB has been used for geometric mapping in robotics applications, semantic mapping for scene understanding with OpenVDB remains unexplored.
In addition, existing semantic mapping systems lack support for integrating both fixed-category and open-language label predictions within a single framework.
In this paper, we propose a novel 3D semantic mapping system that leverages the OpenVDB data structure and integrates a unified Bayesian update framework for both closed- and open-set semantic fusion. 
Our proposed framework, SLIM-VDB, achieves significant reduction in both memory and integration times compared to current state-of-the-art semantic mapping approaches, while maintaining comparable mapping accuracy.
An open-source C++ codebase with a Python interface is available at \url{https://github.com/umfieldrobotics/slim-vdb}.
\end{abstract}

\section{INTRODUCTION}

Robots require accurate world maps with scene understanding upon which to condition their actions.
Recent mapping methods represent both the geometry \cite{vizzo2022} and semantics \cite{zhu2024, ewen2024you} of the robot's surroundings, enabling scene understanding and complex task completion.
A popular approach is to represent a scene using a Truncated Signed Distance Function (TSDF) \cite{curless1996}, which represents the contact surface of the environment and results in realistic dense reconstructions.

However, existing approaches that build TSDF maps have large memory requirements and struggle to run in real time \cite{schmid2022}.
Furthermore, non-probabilistic semantic mapping methods are not robust to inconsistencies or ``flickering" in labels output by semantic segmentation networks \cite{park2022}.
Even as closed-set semantics remain useful for simplified world understanding, the growing use of open-language representations \cite{wilson2024} highlights the need for mapping frameworks that can handle both paradigms within a unified formulation.
These challenges motivate the development of a compute-efficient probabilistic mapping system for mobile robots capable of handling both closed and open-set semantics.

\begin{figure}[ht]
    \centering
    \includegraphics[width=\linewidth]{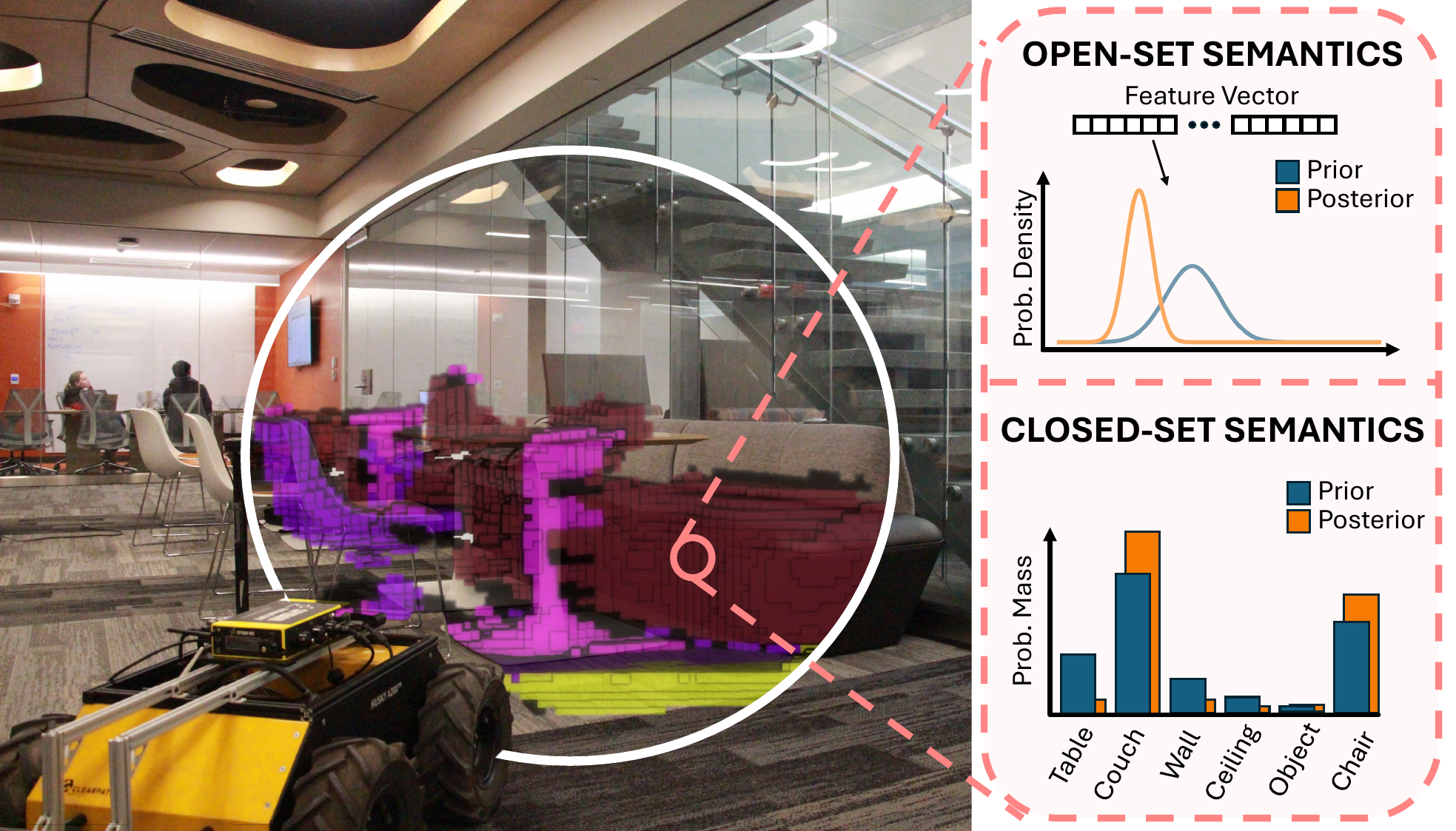}
    \vspace{-1.2em}
    \caption{An example of the real-time semantic mapping method proposed in this paper. SLIM-VDB is capable of using either open- or closed-set semantics and leverages a Bayesian approach for semantic fusion. This approach requires significantly less memory and computation time than existing state-of-the-art semantic mapping methods.}
    \label{fig:cover}
    \vspace{-1.2em}
\end{figure}

The computer graphics community has developed the OpenVDB \cite{museth2013} data structure for efficient volumetric data storage. 
VDBFusion \cite{vizzo2022} has recently adapted OpenVDB into a robotic mapping tool, which demonstrates state-of-the-art performance in computation speed and memory consumption for mapping, while maintaining high geometric accuracy. 
However, VDBFusion lacks semantic mapping capabilities. 

Prior work has used Bayesian updates to deal with either closed-set \cite{ewen2022} or open-set \cite{wilson2024} semantic uncertainty in order to perform semantic mapping.
Unfortunately, these approaches are computationally demanding and are only applicable to either open-set or closed-set semantics, not both.

To address these challenges, we introduce SLIM-VDB, the Semantic Lightweight Implicit Mapping system. SLIM-VDB leverages the OpenVDB data structure and introduces a unified Bayesian semantic fusion update step to enable real-time, memory-efficient semantic mapping for either closed-set or open-set semantics.
As far as we are aware, this is the first work to use an OpenVDB backend for semantic mapping and the first work to present a unified probabilistic semantic fusion update for either open- or closed-set class labels.
We also present an open-source C++ codebase with a Python interface to the greater robotics community. An introduction to the system is shown in Fig. \ref{fig:cover}. 

In summary, the contributions of this paper are as follows:
\begin{itemize}
    \item A novel framework that builds on OpenVDB to enable lightweight, memory-efficient semantic mapping.
    \item A unified Bayesian inference framework that enables either closed-set or open-set semantic estimation.
    \item An open-source C++ library with a Python interface for easy integration with robotics applications.
\end{itemize}

\section{RELATED WORKS}

SLIM-VDB builds off of advances in real-time TSDF mapping and recurrent Bayesian approaches that fuse neural network predictions into a probabilistic map representation. In this section, we discuss related work on geometric and semantic mapping. 

\subsection{Geometric Mapping}

OctoMap \cite{hornung2013} was the first major step towards a real-time, dynamic, multi-resolution, and probabilistic 3D mapping framework, and it has been the standard in the mapping community for some time. 
The emergence of implicit surface representations with TSDFs through Voxblox \cite{oleynikova2017} outperformed OctoMap in update speed and reconstruction quality. 
Voxblox was extended to include instance and semantic information, and Voxfield Panmap \cite{schmid2022, pan2022} has continued improvement in panoptic TSDF voxel mapping. A recent work \cite{vizzo2022} proposed an OpenVDB-based \cite{museth2013} lightweight TSDF mapping framework that outperforms Voxblox in runtime and memory efficiency. This is a promising new direction for efficient volumetric mapping, but there are no current approaches that allow for the incorporation of semantics.

Neural Radiance Fields and 3D Gaussian Splatting \cite{tosi2024} have demonstrated high-quality dense reconstruction but are typically very computationally and memory intensive. Recent work \cite{isaacson2023} has begun to make these methods operate in real-time, however, voxel-based mapping frameworks remain very useful for mobile systems that have limited computational resources.

\subsection{Semantic Mapping}

Maps can store more than just geometric information, such as semantic categories obtained from neural networks. 
While early approaches to semantic mapping employed voting schemes or heuristics \cite{SemanticFrequent3D}, many modern approaches employ learning-based semantic fusion techniques. Semantic Neural Implicit SLAM (SNI-SLAM) \cite{zhu2024} proposes a real-time semantic neural implicit representation that correlates appearance, geometry, and semantic features using cross-attention. However, the system is only able to run in real-time on professional server grade GPUs, which makes it ill-suited for mobile robot platforms.

Probabilistic approaches to semantic fusion  such as Convolutional Bayesian Kernel Inference (ConvBKI) \cite{ConvBKI2} and SEE-CSOM \cite{deng2023} recurrently incorporate semantic predictions into maps through Bayesian inference.
One benefit of Bayesian inference is the quantification of uncertainty, which can be leveraged for downstream applications such as active perception. 
By adopting a conjugate prior over a parametric likelihood function that models the distribution of the data, the expectation and variance of voxel-wise semantics can be monitored and updated over time as new measurements are obtained \cite{MappingSBKI, ewen2022, ewen2024you, ConvBKI2}. 
However, closed-set semantic maps are limited by a fixed number of categories, which may not be suitable for more complex environments or out-of-distribution data.

Recently, the deep learning community has developed Vision-Language Models (VLMs), which generate predictions in a latent space shared with language models, enabling open-dictionary language-based queries \cite{radford2021}. Instead of requiring a fixed set of categories at model selection, open-dictionary segmentation models enable categorization at inference, which allows for reasoning about synonymous categories or obscure objects. 
LatentBKI \cite{wilson2024} extends open-dictionary semantics to mapping by adopting a conjugate prior over a likelihood function that describes the distribution of the feature embeddings. However, it does not run in real-time, and the map size is greatly limited by GPU memory. 

\begin{figure*}[ht]
    \centering
    \includegraphics[width=0.99\linewidth]{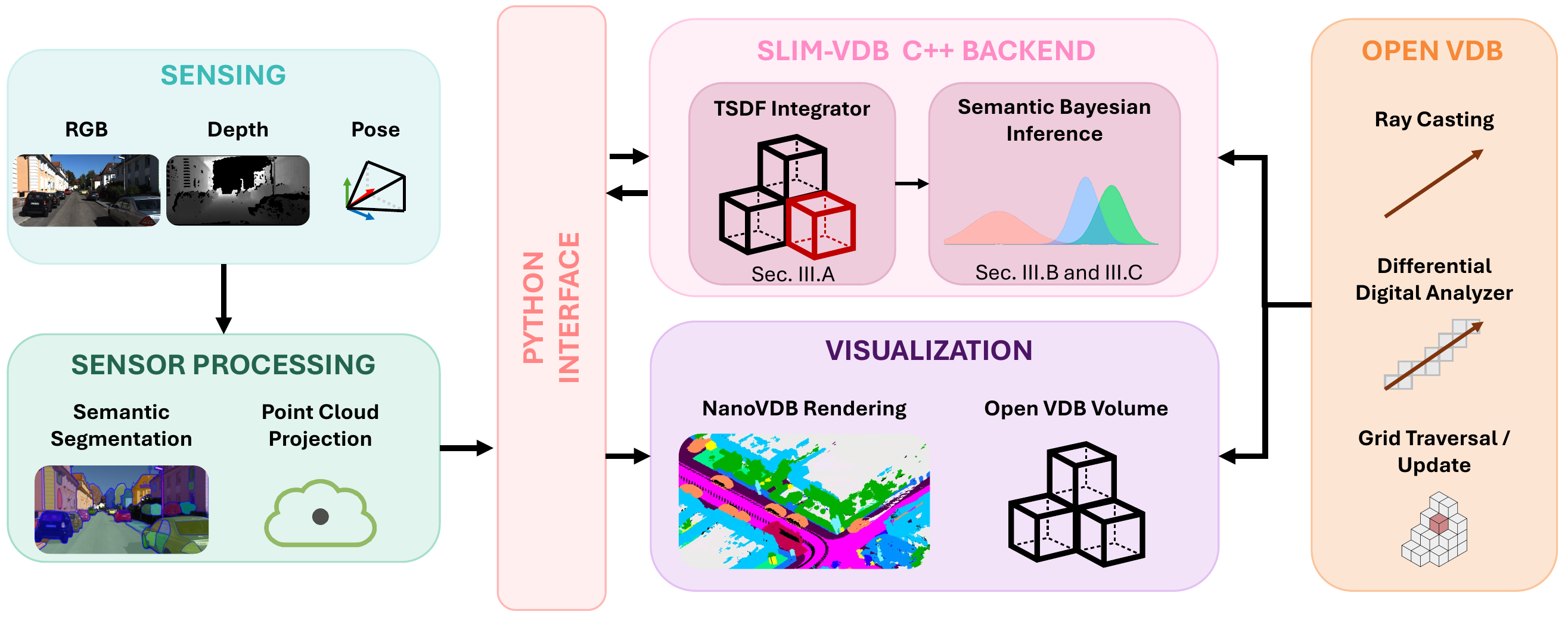}
    \vspace{-1em}
    \caption{Overview of the SLIM-VDB mapping system. The input dataset can be either a pointcloud or RGB-D images. Semantic segmentation can be performed with any off-the-shelf network, and we use Semantic Segment Anything \cite{chen2023}. The main integration loop conducts raycasting and TSDF updates (\ref{subsec:geometry_update}) as well as a probabilistic update for either closed-set (\ref{subsec:closed_set}) or open-set (\ref{subsec:open_set}) semantics. Finally, the rendering (\ref{subsec:rendering}) utilizes NanoVDB \cite{museth2021} for GPU-accelerated viewing of the map.}
    \label{fig:flowchart}
    \vspace{-1em}
\end{figure*}

\subsection{OpenVDB}
OpenVDB \cite{museth2013} was developed at DreamWorks Animation to address a critical need for efficient volumetric mapping of animated character assets. 
The key strengths of OpenVDB lie in its $\mathcal{O}(1)$ lookup, insertion, and delete times, along with a low memory footprint due to its use of a B+ tree to limit the number of levels in the tree.
OpenVDB was a pioneer in the use of compact ``level set'' data structures, which is otherwise known as signed distance fields to the robotics and computer vision community. 

OpenVDB has begun to be repurposed for highly efficient volumetric map data storage for mobile robots. 
The use of OpenVDB led to various efficiency improvements in point insertion and raycasting \cite{besselman2021, hagmanns2022}, distributed systems \cite{besselman2022}, and 3D reconstruction quality \cite{bai2023}. VDBFusion \cite{vizzo2022} is a open-source library that takes advantage of OpenVDB's robust C++ implementation and is capable of integrating LiDAR data at 20 frames per second into a map, outperforming Voxblox and OctoMap. However, VDBFusion does not support the incorporation of semantic information into the map, nor does it utilize NanoVDB \cite{museth2021} -- a recent GPU-based improvement to OpenVDB -- for real-time rendering. The only work that has utilized NanoVDB for map rendering, NanoMap \cite{walker2022}, also does not incorporate semantics.

In this work, we develop a GPU-accelerated OpenVDB-based semantic mapping framework for robotics applications that runs in real-time and with a lower memory footprint than existing state-of-the-art approaches. 
Our framework uses a unified Bayesian probabilistic update paradigm for semantic label fusion that allows for the use of either closed-set \cite{ewen2022} or open-set \cite{wilson2024} semantic information. 

\section{METHODOLOGY}

This section presents the proposed semantic mapping framework, illustrated in Fig.~\ref{fig:flowchart}.
First, semantic information, either using open- or closed-set semantics, is computed for images collected using a robot's onboard RGB-D camera or LiDAR.
These predictions are projected using the depth image into a semantic point cloud.
The geometric information of this point cloud is used to update the TSDF map as described in Section \ref{subsec:geometry_update}.
Next, the global semantic information is updated using Bayesian inference for either closed-set (Section \ref{subsec:closed_set}) or open-set (Section \ref{subsec:open_set}) semantics.
Lastly, the rendering pipeline is presented to visualize the semantic TSDF in Section \ref{subsec:rendering}. 

Vectors, written as columns, are typeset in bold, sets are typeset in uppercase calligraphic font and random variables are uppercase. 
The element $i$ of a vector $\mathbf{x}$ is denoted as $x_i$.
Let $p$ denote a normalized probability mass or density function. 
We use $_*$ to indicate operations at the voxel level.

\subsection{Surface Reconstruction} \label{subsec:geometry_update}

The proposed semantic mapping pipeline stores the geometric information of a scene in voxels using the OpenVDB library.
This library enables fast and memory-efficient storage and querying via a sparse, hierarchical data structure that is used to store the signed distance, weight, and semantic parameters of voxels close to a surface.

At each frame, a semantically segmentated image is projected to 3D using a corresponding depth image. This semantic point cloud is used to sequentially update the voxel map.
Raycasting, which determines which voxels correspond to free space and which must be updated, is the critical and most computationally costly operation of this process.
We adopt \cite{vizzo2022}'s implementation of raycasting, which employs OpenVDB's Differential Digital Analyzer \cite{museth2013}.
Motivated by \cite{curless1996}, the raycasting method considers only voxels in a configurable ``truncated'' region around each point, minimizing the integration step computational cost.

Following \cite{curless1996} for volumetric integration with a signed distance function representation, the truncated signed distance and voxel weight values are incrementally updated as new point clouds are integrated into the map.
That is, for voxel $\mathbf{x}_* \in \mathbb{R}^3$ at time $t$, the cumulative truncated signed distance, $D_t(\mathbf{x}_*)$, and cumulative weight, $W_t(\mathbf{x}_*)$, of the voxel are computed using the signed distance from the latest sensor measurement, $d_t(\mathbf{x}_*)$, and the weighting function, $\Tilde{w}(\mathbf{x}_*)$:
\begin{equation}
    D_t(\mathbf{x}_*) = \frac{W_{t-1}(\mathbf{x}_*) \; D_{t-1}(\mathbf{x}_*) + \tilde{w}(\mathbf{x}_*) \; d_t(\mathbf{x}_*)}{W_t(\mathbf{x}_*)},
\end{equation}
\begin{equation} \label{eq:Wx}
    W_t(\mathbf{x}_*) = W_{t-1}(\mathbf{x}_*) + \Tilde{w}(\mathbf{x}_*).
\end{equation}
\noindent A popular choice for a weighting function is $\Tilde{w}(\mathbf{x}_*) = 1$.

\subsection{Unified Bayesian Semantic Fusion}

In our unified approach to closed or open-set semantic fusion, we utilize a closed-form conjugate pair recursive Bayesian update to estimate the semantic class at each voxel. In probability theory, conjugacy refers to a pair of distributions where, given a prior distribution and a likelihood function, the posterior belongs to the same family of distributions as the prior \cite{tu2014, murphy2007conjugate}.
This results in a tractable solution for the posterior where only the hyperparameters of the prior need to be updated to reflect the new information from the measurement. Due to the differing likelihood distributions, the Dirichlet-Categorical conjugacy is used for closed-set semantics and the Normal Inverse Gamma-Normal conjugacy is used for open-set semantics.

\subsection{Closed-Set Semantic Fusion} \label{subsec:closed_set}

We use the Dirichlet-Categorical conjugate pair for closed-set semantic fusion as also shown in \cite{MappingSBKI, ewen2022}.
Let $z \in  \{1,\ldots,k\}$ denote a discrete Categorical random variable in the predefined sample space with the probability mass function:
\begin{equation} \label{eq:categorical}
    p(z = i \mid \boldsymbol{\theta}) = \text{Cat}(z = i \mid \boldsymbol{\theta}) = \theta_i,
\end{equation}
where $i \in \{1, 2, \dots, k\}$, $\boldsymbol{\theta} \in [0, 1]^k$, and $\sum^k_{i=1} \theta_i = 1$.
Given the parameter $\boldsymbol{\theta}$, this represents the probability that a sample $z$ belongs to semantic class $i$.

The conjugate prior for the Categorical distribution is the Dirichlet distribution.
The Dirichlet distribution is the multivariate generalization of the beta distribution that is parameterized by a vector $\boldsymbol{\alpha} \in \mathbb{R}^k_{\geq 0}$ and has the probability density: 
\begin{equation} \label{eq:dirichlet}
    p(\boldsymbol{\theta} \mid \boldsymbol{\alpha}) = Dir(\boldsymbol{\theta} \mid \boldsymbol{\alpha})
    = \frac{\Gamma(\sum_{j=1}^k \alpha_j)}{\sum_{j=1}^k \Gamma(\alpha_j)} \prod_{j=1}^k \theta_j^{\alpha_j-1},
\end{equation}
where $\Gamma(\alpha_j)$ is the Gamma function.
The conjugacy of these two distributions is formalized in the following theorem \cite{ewen2022}:

\begin{thm} \label{thm:dir_cat_conj}
Consider a set of semantic observations $\mathcal{Z} = \{z_1, \dots, z_n\}$, each independently drawn from a Categorical distribution $p(\mathcal{Z} \mid  \boldsymbol{\theta})$. Let the prior be a Dirichlet distribution, $p(\boldsymbol{\theta} \mid \boldsymbol{\alpha})$. Then, the posterior of $\boldsymbol{\theta}$ given the observations $\gZ$ and parameters $\boldsymbol{\alpha}$ is also a Dirichlet distribution:
\begin{equation} \label{eq:dir_posterior}
    p(\boldsymbol{\theta} \mid \gZ, \boldsymbol{\alpha}) = Dir(\boldsymbol{\theta} \mid \Tilde{\boldsymbol{\alpha}}),
\end{equation}
where the updated Dirichlet parameters are given by:
\begin{align} \label{eq:alpha_update}
    \tilde{\alpha}_i &= \alpha_i + \sum_{z_j \in \gZ} \mathds{1}\{z_j = i\},
\end{align}
and $\mathds{1}\{z_j = i\}$ is an indicator function that equals 1 if observation $z_j$ belongs to class $i$, and 0 otherwise.
\end{thm}

Leveraging Theorem \ref{thm:dir_cat_conj} enables us to take a Bayesian approach to integrating multiple semantic predictions at one voxel over time. 
At each timestep $t$, a set of 3D points and their corresponding closed-set semantic predictions, $\mathcal{P}_C = \{(\mathbf{x}_j, Z_j)\}_{j=1}^N$ where $\mathbf{x}_j \in \mathbb{R}^3$ and $Z_j \in \{1,\dots, k\}$ is a discrete random variable, is derived from either an RGB-D image or a point cloud.
This semantic prediction is typically represented as a one-hot vector output of a segmentation network. 
At each voxel, Bayes' Rule is recurrently applied to compute the Dirichlet posterior distribution using Eq. \eqref{eq:alpha_update}. 

In general, computing the predictive posterior involves marginalizing over $\boldsymbol{\theta}$, which can be intractable. However, due to the Dirichlet-Categorical conjugate pair there exists a closed-form solution \cite{tu2014, ewen2022}:
\begin{equation} \label{eq:dir_to_cat}
    p(z = i\mid \tilde{\boldsymbol{\alpha}}) = \frac{\tilde{\alpha}_i}{\sum_{j=1}^k \tilde{\alpha}_j}.
\end{equation}

This defines the posterior predictive distribution over the closed-set semantic class. In practice, it is sufficient to track the Dirichlet parameter $\boldsymbol{\alpha}$ to compute these predictions. We will explore in the next section how a similar Bayesian framework can also handle open-set semantics from VLMs.

\subsection{Open-Set Semantic Fusion} \label{subsec:open_set}

Motivated by the use of conjugate pairs for Bayesian inference from the previous section, we use the Normal Inverse Gamma and Normal conjugate pair for open-set semantic fusion as introduced in \cite{wilson2024}.

Let $\mathbf{Z} \in \mathbb{R}^l$ denote a multivariate, continuous Gaussian random variable with independent elements, where each element has the probability density:
\begin{align}
    p(Z_i \mid \mu_i, \sigma_i) &= \mathcal{N}(Z_i \mid \mu_i, \sigma_i^2).
\end{align}

The Normal Inverse Gamma distribution is a multivariate continuous probability distribution parameterized by four variables: $m$, $\lambda$, $\nu$, and $\beta$.
Together, these parameters define a probability density for the mean, $\mu$, and variance, $\sigma^2$, of the Normal distribution such that:
\begin{align}
p(\mu_i, \sigma_i^2 \mid m_i, \lambda_i, \nu_i, \beta_i) &= \mathcal{N}(\mu_i \mid m_, \sigma_i^2 / \lambda_i) \\
& \;\;\;\;\;\;\;\;\;\;\;\;\; \cdot \Gamma^{-1}(\sigma_i^2 \mid \nu_i, \beta_i),
\end{align}
where $\Gamma^{-1}$ is the inverse gamma function. The Normal Inverse Gamma distribution is the conjugate pair to the Normal distribution, as is formalized in the following theorem:

\begin{thm} \label{thm:norm_inv_conj}
Let $\gZ_i = \{Z_{i,1}, \dots, Z_{i,n}\}$ be a set of samples drawn from the same Normal distribution, $p(Z_{i,j} \mid \mu_i, \sigma_i^2)$, with unknown mean and variance and let the prior for $\mu_i, \sigma_i^2$ be a Normal Inverse Gamma distribution, $p(\mu_i, \sigma_i^2 \mid m_i, \lambda_i, \nu_i, \beta_i)$.
The posterior computed using Bayes' theorem, $p(\mu_i, \sigma_i^2 \mid \gZ_i, m_i, \lambda_i, \nu_i, \beta_i)$, is also a Normal Inverse Gamma distribution.
The closed form equations to compute the parameters are as follows, where $|\gZ_i|$ is the cardinality of $\gZ_i$ and $\bar{\gZ_i}$ is the mean of $\gZ_i$:
\begin{equation}\label{eq:norm_update_1}
    \Tilde{m}_i = \frac{\lambda_i m_i + |\gZ_i| \bar{\gZ_i}}{\Tilde{\lambda}_i},
\end{equation}
\begin{equation}
    \Tilde{\lambda}_i = \lambda_i + |\gZ_i|,
\end{equation}
\begin{equation}
    \Tilde{\nu}_i = \nu_i + \frac{|\gZ_i|}{2},
\end{equation}
\begin{equation}\label{eq:norm_update_4}
    \scalebox{0.95}[1]{$\Tilde{\beta}_i = \beta_i + \frac{1}{2}\sum_{z_{i,j} \in \gZ_i} (z_{i,j} - \bar{\gZ_i})^2 + \frac{\lambda_i |\gZ_i|}{\Tilde{\lambda}_i} \frac{(\bar{\gZ_i} - m_i)^2}{2}$}.
\end{equation}\vspace{0.1em}
\end{thm}

For open-set semantics, we model the measurement likelihood in the VLM feature space as a multivariate Gaussian random variable with i.i.d. elements.
Given an RGB-D image and a VLM, we derive a set of 3D points and their corresponding open-set semantic predictions, $\mathcal{P}_O = \{(\mathbf{x}_i, \mathbf{Z}_i)\}_{i=1}^N$ where $\mathbf{x}_i \in \mathbb{R}^3$ is the 3D point and $\mathbf{Z}_i \in \mathbb{R}^l$ is the aforementioned multivariate continuous Gaussian random variable.
Notably, the mean and variance of each element of $\mathbf{Z}_i$ is unknown.
This enables us to represent the feature embedding as a vector of $l$ 1-dimensional Gaussian random variables, each with an unknown mean and variance.

Following Theorem \ref{thm:norm_inv_conj}, we utilize the conjugacy between the Normal distribution and the Normal Inverse Gamma to tractably compute the posterior for each element of $\mathbf{Z}_i$. This is done through the parameter updates defined in Eqs. \eqref{eq:norm_update_1}-\eqref{eq:norm_update_4}. Theorem \ref{thm:norm_inv_conj} then enables the estimate of the open-set semantics via the predictive posterior for the Gaussian random variable $Z_i$ given previous semantic measurements:
\begin{equation} \label{eq:student_t}
    \scalebox{0.99}[1]{$p(Z_i \mid \gZ_i, m_i, \lambda_i, \nu_i, \beta_i) = t_{2\nu_i}\left(Z_i \mid m_i, \frac{\beta_i (\lambda_i + 1)}{\lambda_i \nu_i}  \right)$}
\end{equation}
\noindent where $t_{2\nu_i}$ is the Student-$t$ distribution \cite{murphy2007conjugate}. The expectation of the Student-$t$ distribution is $\mathbb{E}(Z_i) = m_i$ when $\lambda_i > 1$, and a covariance of $\text{Cov}(Z_i) = \frac{\lambda_i}{\lambda_i - 2} \left( \frac{\beta_i(\lambda_i + 1)}{\lambda_i \nu_i} \right)$ for $\lambda_i > 2$. This results in a distribution over the feature vectors, and we then apply a softmax in order to get absolute class probabilities.

\begin{table*}[!htb]
\centering
\setlength{\defaultaddspace}{2pt}
\caption{Comparison of average FPS and CPU+GPU runtime memory usage for closed- $\left(^\text{C}\right)$ and open-set $\left(^\text{O}\right)$ semantic mapping, including $\pm$ one std. deviation. SLIM-VDB* denotes a heuristic version of our method with a non-Bayesian semantic update. \textbf{Bold} is best.}
\label{table:memory_runtime_closed_set}
\begin{tabular}{clcccccc}
\toprule
    &&& \multicolumn{2}{c}{SemanticKITTI \cite{behley2019}} & & \multicolumn{2}{c}{SceneNet \cite{mccormac2017}} \\
    & Method && FPS $\uparrow$ & Memory (GB) $\downarrow$ & & FPS $\uparrow$ & Memory (GB) $\downarrow$ \\
\midrule
\midrule
\multirow{3}{*}{Non-Bayesian} & SNI SLAM \cite{deng2023} && 0.13 $\pm$ 0.02 & 19.43 $\pm$ 0.98 && 0.27 $\pm$ 0.01 & 19.43 $\pm$ 0.07 \\
                                & VoxField \cite{schmid2022} && 4.43 $\pm$ 0.72 & 3.06 $\pm$ 0.63 && 2.49 $\pm$ 1.67 & 1.05 $\pm$ 0.45 \\
                              & SLIM-VDB* (Ours) && \textbf{6.25 $\pm$ 1.12} & \textbf{0.67 $\pm$ 0.05} && \textbf{16.55 $\pm$ 2.87} & \textbf{0.51 $\pm$ 0.01} \\
\midrule
\multirow{4}{*}{Bayesian} & ConvBKI \cite{wilson2024} && 0.53 $\pm$ 0.26 & 27.73 $\pm$ 0.20 && 1.10 $\pm$ 1.00 & 14.32 $\pm$ 7.02 \\
                              & SEE-CSOM \cite{deng2023} && 0.10 $\pm$ 0.01 & 10.98 $\pm$ 0.35 && 2.67 $\pm$ 8.00 & 5.19 $\pm$ 1.30 \\
                              & SLIM-VDB$^\text{C}$ (Ours) && \textbf{1.69 $\pm$ 0.69} & \textbf{2.60 $\pm$ 0.80} && \textbf{10.84 $\pm$ 1.94} & \textbf{1.08 $\pm$ 0.16} \\
\addlinespace
\cdashline{2-8}
\addlinespace
& LatentBKI \cite{wilson2024} && - & - && 1.67 $\pm$ 0.36 & 27.68 $\pm$ 1.37 \\
& SLIM-VDB$^\text{O}$ (Ours) && - & - && \textbf{1.76 $\pm$ 0.11} & \textbf{3.49 $\pm$ 0.68} \\
\bottomrule
\end{tabular}
\end{table*}

\begin{figure*}[ht]
    \centering
    \includegraphics[width=0.915\linewidth]{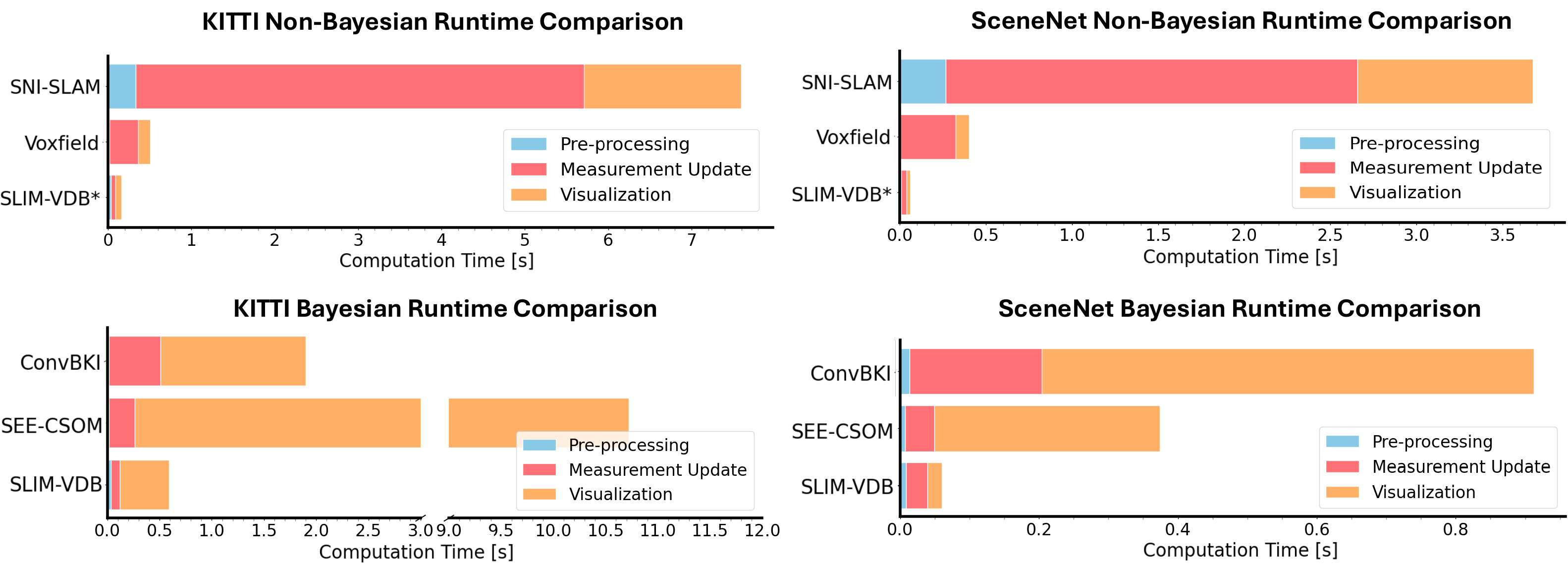}
    \vspace{-1em}
    \caption{Visual runtime comparison on SceneNet and KITTI scenes for both Bayesian and non-Bayesian semantic mapping frameworks.}
    \label{fig:comp_times}
    \vspace{-1em}
\end{figure*}

In the implementation, we store the mean vector $\mathbf{m}_* = (\Tilde{m}_1, \Tilde{m}_2, \dots, \Tilde{m}_l)$ at a voxel level, where $l$ is the size of the feature vector. We simplify the recursive estimation of the posterior by using the cumulative TSDF weight $W_t(\mathbf{x}_*)$ for $\lambda_*$ and $2 \nu_*$. Additionally, due to our independence assumption for each element of the feature vector, we only track the diagonal of the covariance matrix $\beta$. During inference, the highest class probability at each voxel resulting from the posterior predictive is thresholded against $T_p = 0.1$, in order to filter out uncertain class predictions.

\subsection{Rendering} \label{subsec:rendering}

The truncated signed distance values, weight, and semantic parameters corresponding to either open- or closed-set semantics are stored in voxels contained inside the OpenVDB data structure.
Rather than constructing a voxel mesh and publishing it to a visualizer such as Rviz, we use the native VDB rendering tools to provide real-time views of the scene, which render the grid from a given camera viewpoint. 
Rendering is traditionally a computationally intensive operation, so we utilize NanoVDB~\cite{museth2021}, a GPU-optimized version of OpenVDB, for faster, parallelized rendering.

\section{EXPERIMENTS}

We validate the proposed SLIM-VDB algorithm on both simulation and real world benchmarks, demonstrating a significant reduction in computation time and memory consumption while maintaining comparable accuracy to state-of-the-art semantic mapping approaches. All experiments except those in subsection \ref{sec:mob-hw} were run on an Intel Xeon W-2245 CPU @ 3.90GHz with an NVIDIA GeForce 32 GB RTX 3090 GPU. All experiments on KITTI \cite{behley2019} map at a 10 cm per voxel resolution, closed-set experiments on SceneNet \cite{mccormac2017} map at a 5 cm per voxel, and open-set experiments on SceneNet map at 10 cm per voxel. Evaluation is run in C++.

\subsection{Closed-Set Semantic Mapping}
We compare SLIM-VDB against five state-of-the-art semantic mapping approaches: ConvBKI \cite{ConvBKI2}, SEE-CSOM \cite{deng2023}, Voxfield Panmap \cite{schmid2022}, SNI-SLAM \cite{zhu2024}, and LatentBKI \cite{wilson2024}.
We evaluate runtime and memory consumption given ground truth semantic labels. For semantic prediction accuracy, we use segmentation results from the Semantic Segment Anything \cite{chen2023} network for SceneNet, and PolarSeg \cite{zhou2021} for SemanticKITTI. PolarSeg is finetuned on SemanticKITTI scenes 0-5, and we evaluate on the first 100 frames of scenes 6-10. We randomly select 12 full SceneNet scenes for evaluation from two different trajectories.

SemanticKITTI \cite{behley2019} and SceneNet \cite{mccormac2017} are used to evaluate SLIM-VDB on large-scale scenes and on indoor simulated scenes, respectively.
ConvBKI and SEE-CSOM both include a Bayesian update for closed-set semantic Dirichlet parameters (see Section \ref{subsec:closed_set}). 
In order to evaluate ConvBKI we do not use free space sampling, since the global mapping approach does not scale with voxels introduced from free space. We would like to note that this is a significant advantage of our method SLIM-VDB, as it employs a similar Bayesian update with improved global scalability. 

In contrast, Voxfield Panmap and SNI-SLAM assume consistent and error-free semantic predictions.
As such, they do not use Bayesian inference for semantic estimation.
For a fair comparison, we evaluate our approach without the Bayesian update step -- denoted as SLIM-VDB* in the results -- against Voxfield Panmap and SNI-SLAM.





\begin{figure*}[!ht]
    \centering
    \includegraphics[width=0.955\linewidth]{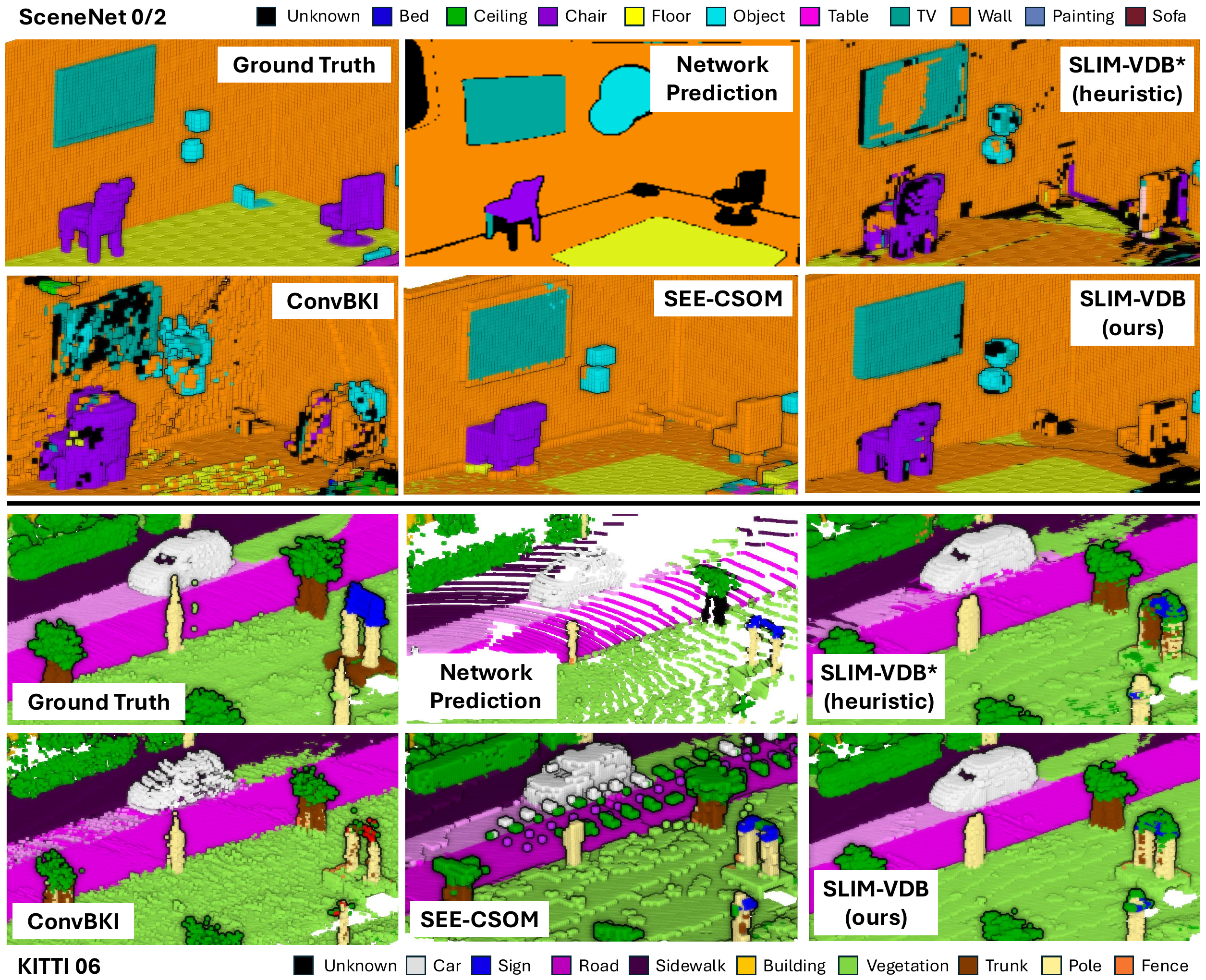}
    \caption{Qualitative comparison on SceneNet 0/2 and SemanticKITTI 06 between the ground truth, the network output on a single frame, SLIM-VDB, and other probabilistic baselines (SEE-CSOM \cite{deng2023} and ConvBKI \cite{ConvBKI2}).}
    \label{fig:qual_comp}
    \vspace{-1em}
\end{figure*}

\subsubsection{Runtime}

Table \ref{table:memory_runtime_closed_set} presents the average frames per second (FPS) for each method. SLIM-VDB has the shortest runtime on both the large-scale outdoor SemanticKITTI and indoor SceneNet scenes, outperforming all probabilistic baselines with the highest FPS.
When the Bayesian update is not used, SLIM-VDB* also outperforms the non-probabilistic baselines Voxfield Panmap and SNI-SLAM.
This demonstrates the efficiency improvements of having OpenVDB as a mapping backbone. Fig. \ref{fig:comp_times} visualizes the average runtimes for the pre-processing, integration, and visualization steps of a single frame across all methods.
SLIM-VDB outperforms all baseline methods for each step with notable improvements to the integration. Although SLIM-VDB also greatly improves on baseline visualization cost, these are less comparable because our method only renders a single viewpoint of the scene at every frame rather than publishing a mesh to Rviz.


\subsubsection{Memory}

SLIM-VDB also greatly outperforms all baseline methods with respect to combined CPU and GPU memory consumption, as shown in Table \ref{table:memory_runtime_closed_set}. This reduction in memory consumption is due to the underlying properties of the OpenVDB structure, which does not require initializing all voxels within the map bounds. In contrast, ConvBKI recurrently updates the map through a dense convolutional filter, which requires the entire local map to be stored on GPU memory. Additionally, SLIM-VDB does not require the user to set a static global map size beforehand, as required by ConvBKI and SEE-CSOM.

\subsubsection{Semantic Prediction Accuracy}

Finally, we compare the accuracy of SLIM-VDB against the closed-set probabilistic mapping baselines, ConvBKI and SEE-CSOM. In Table \ref{table:miou}, we present averaged mean Intersection over Union (mIoU) across 12 random SceneNet scenes and five SemanticKITTI scenes. Qualitative results are shown in Fig. \ref{fig:qual_comp}. For SceneNet, SLIM-VDB and SEE-CSOM have comparable performance and both significantly outperform ConvBKI. For the KITTI dataset, SLIM-VDB$^\text{C}$ outperforms all baselines, including SEE-CSOM. This is likely due to geometric errors in the SEE-CSOM results with larger maps, as discussed in \ref{sec:geom}. We include a non-probabilistic ablation SLIM-VDB*, which simply updates the semantic prediction at each voxel according to the most recent measurement, demonstrating the value of our probabilistic update.
Note that a voxel-wise comparison rewards geometric accuracy and penalizes filling in extra voxels.

\begin{table}[!htb]
    \setlength{\tabcolsep}{4pt} 
    \centering
    \caption{Avg semantic accuracy results on SceneNet (SN) and KITTI.}
    \label{table:miou}
    \begin{tabular}{lcccc:c}
        \toprule
        & \multicolumn{4}{c}{mIoU $\uparrow$} \\ 
        & \rotatebox{45}{ConvBKI} & \rotatebox{45}{SEE-CSOM} & \rotatebox{45}{SLIM-VDB*} & \rotatebox{45}{SLIM-VDB$^\text{C}$} & \rotatebox{45}{SLIM-VDB$^\text{O}$} \\
        \midrule
        \midrule
        KITTI & 0.165 & 0.128 & \underline{0.216} & \textbf{0.252} & - \\
        \midrule
        SN & 0.049 & \textbf{0.128} & 0.087 & \underline{0.122} & 0.116 \\
        \bottomrule
    \end{tabular}
    \vspace{-0.8em}
\end{table}

\subsection{Open-Set Semantic Mapping}

In this section, we demonstrate the application of a Bayesian semantic probabilistic update framework to open-set semantics using the Normal Inverse Gamma distribution.
For the closed-set semantics evaluation, we use Semantic Segment Anything \cite{chen2023} to generate open-set text labels for each pixel in the SceneNet trajectories and then bin the text labels into the pre-defined SceneNet categories. For fair comparison, we use the same open-set text labels previously generated for the closed-set experiment and then embed them into the feature space of size $\mathbb{R}^{512}$ with Contrastive Language-Image Pre-training (CLIP) \cite{radford2021}. These implicit semantic feature vectors are then integrated into the volume at every time step using the Normal Inverse Gamma update as previously described.

\subsubsection{Runtime}

We indicate the runtime performance of SLIM-VDB with open-set labels (denoted by SLIM-VDB$^\text{O}$) in Table \ref{table:memory_runtime_closed_set}. Note that the runtime reflects the average pre-processing and integration times across the 300 frames of the SceneNet scene, but does not include visualization time. This is because LatentBKI has no real-time visualization.


\subsubsection{Memory}

The memory performance (combined CPU and GPU) of SLIM-VDB with open-set labels is also shown in Table \ref{table:memory_runtime_closed_set}. The use of neural implicit representations in the baseline LatentBKI leads to a very large GPU memory usage. We evaluated the two open-set methods at 10 cm per voxel resolution rather than 5 cm per voxel resolution used for the closed-set experiments as the testbench did not have enough memory to evaluate LatentBKI with a smaller resolution.

\subsubsection{Semantic Prediction Accuracy}

\begin{figure}[!ht]
    \centering
    \includegraphics[width=\linewidth]{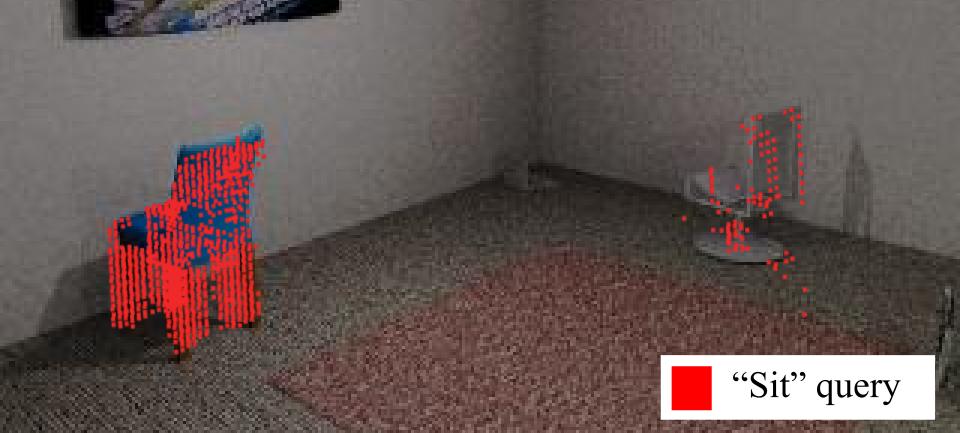}
    \vspace{-1.5em}
    \caption{Ground truth SceneNet 0/2 RGB image with overlayed red voxels showing regions with SLIM-VDB feature vectors of a high similarity to a text embedding for ``sit.'' We notice that chairs are highlighted as potential places to take a ``sit'' action.}
    \label{fig:sit}
    \vspace{-1.5em}
\end{figure}

To evaluate, we show a comparison against the closed-set (SLIM-VDB$^\text{C}$) mapping results in Table \ref{table:miou}. The open-set and closed-set SLIM-VDB semantic accuracy is very comparable, reinforcing the unified Bayesian update framework proposed in this work. Some potential users of this system may prefer the expressiveness of open-set semantics, as the semantic class ``chair" could be more directly linked to the downstream action ``sit." See Fig. \ref{fig:sit} for an overlay of SLIM-VDB voxels with semantics similar to the CLIP embedding for ``sit'' in red. Similarity in the feature space is measured using cosine similarity when querying against a single word or phrase.

\subsection{Geometric Mapping Accuracy}\label{sec:geom}

In this experiment we demonstrate that the volumetric mapping accuracy of SLIM-VDB is comparable to our baselines (see Table \ref{table:chamfer}). We use the L2 Chamfer distance on the final exported map from our selected SceneNet trajectories. SLIM-VDB and VDBFusion have identical geometric mapping accuracy, as they both have the same underlying TSDF update framework. Our method is comparable in accuracy to SEE-CSOM in the indoor SceneNet scenes, although it is slightly outperformed due to the SEE-CSOM method's approach to removing less certain voxels using an entropy heuristic. However, SLIM-VDB drastically outperforms all methods in the outdoor scene as the baseline methods require more memory to map the entire scene than the testbench can provide, and SEE-CSOM's entropy heuristic results in ``boxy" objects not representative of the real scene. ConvBKI populates extra voxels into the map due to noisy depth measurements without free space sampling. We are unable to include geometric accuracy results from SNI-SLAM as the NeRF sampling step is too computationally expensive for our machines.


\begin{table}[!ht]
    \setlength{\tabcolsep}{4pt}
    \centering
    \caption{Geometric accuracy of resulting maps. \textbf{Bold} is best.}
    \label{table:chamfer}
    \vspace{-0.5em}
    \begin{tabular}{lcccccc}
        \toprule
        & \multicolumn{5}{c}{Average Chamfer Distance ($\downarrow$)} \vspace{0.2em} \\
        & VoxField & SNI-SLAM & ConvBKI & SEE-CSOM & SLIM-VDB \\
        \midrule
        \midrule
        KITTI & 6.079 & - & 4.000 & 504.952 & \textbf{0.079} \\
        SN & 0.008 & - & 1.189 & \textbf{0.003} & 0.004 \\
        \bottomrule
    \end{tabular}
    \vspace{-1em}
\end{table}

\subsection{Rendering}

We compare rendering from the VDB map with both CPU and GPU. Over 10 images of size 691 x 256, the average GPU render speed is 2.1 ms and the average CPU render speed is 181.8 ms. For a larger image of 1209 x 448 px, the GPU average render speed is 3.9 ms and the CPU average render speed is 527.7 ms.
If the user does not have access to a GPU, the mapping program could be run headless, as the map pre-processing and integration is run only on the CPU.

\subsection{Mobile Robot Hardware}
\label{sec:mob-hw}

We present experiments on an NVIDIA Jetson Orin Development Kit with just 4GB of GPU memory to demonstrate the usability of SLIM-VDB on real mobile robot hardware. We run SLIM-VDB with closed-set semantics, visualization, and a resolution of 5 cm per voxel on the selected SceneNet scenes, and the resulting average FPS is 6.12. This demonstrates that running SLIM-VDB on mobile robot hardware on indoor scenes is feasible.

\section{CONCLUSIONS}

In this work, we present SLIM-VDB: a 3D probabilistic semantic mapping framework that leverages OpenVDB's efficient volumetric data storage. Our proposed system supports closed- or open-set semantics in a single parallel Bayesian update framework, which to the best of our knowledge is the first single mapping package to do so.
In our experiments, we find that SLIM-VDB is able to perform integration of new data points into the map at higher speeds and with a lower memory consumption than the baselines, while retaining comparable semantic and geometric accuracy.
Additionally, we integrate NanoVDB for GPU-accelerated rendering.
One limitation of our approach is that the real-time viewing of the 3D map is limited to whichever camera rendering viewpoints are previously chosen. We hope to address this in the future by adapting the existing NanoVDB Viewer tool to update dynamically during mapping.
Additionally, future work will include integration with fVDB \cite{williams2024}, a new deep learning toolset that enables convolution and splatting on VDB grids. In order to propel the usability of SLIM-VDB, an open-source codebase will be released.




\section*{ACKNOWLEDGMENT}

This work was supported in part by NSF Grant No. DGE 2241144 and in part by AFOSR MURI under Grant FA9550-23-1-0400.


\balance

\bibliographystyle{ieeetr}
\bibliography{references}

@InProceedings{zhu2024,
    author    = {Zhu, Siting and Wang, Guangming and Blum, Hermann and Liu, Jiuming and Song, Liang and Pollefeys, Marc and Wang, Hesheng},
    title     = {\href{https://arxiv.org/pdf/2311.11016}{{SNI-SLAM}: Semantic Neural Implicit {SLAM}}},
    booktitle = {Proceedings of the IEEE/CVF Conference on Computer Vision and Pattern Recognition (CVPR)},
    month     = {June},
    year      = {2024}
}

@ARTICLE{ewen2022,
  author={Ewen, Parker and Li, Adam and Chen, Yuxin and Hong, Steven and Vasudevan, Ram},
  journal={IEEE Robotics and Automation Letters}, 
  title={\href{https://ieeexplore.ieee.org/document/9792203}{These Maps are Made for Walking: Real-Time Terrain Property Estimation for Mobile Robots}}, 
  year={2022},
  volume={7},
  number={3},
  pages={7083-7090}
}

@article{hornung2013,
  author={Hornung, Armin and Wurm, Kai M. and Bennewitz, Maren and Stachnizz, Cyrill and Burgard, Wolfram},
  journal={Autonomous Robots},
  title={\href{https://dl.acm.org/doi/10.1007/s10514-012-9321-0}{Octo{M}ap: An efficient probabilistic {3D} mapping framework based on octrees}},
  year={2013},
  volume={34},
  pages={189-206}
}

@inproceedings{oleynikova2017,
  title={\href{https://ieeexplore.ieee.org/document/8202315}{Voxblox: {I}ncremental 3{D} {E}uclidean signed distance fields for on-board {MAV} planning}},
  author={Oleynikova, Helen and Taylor, Zachary and Fehr, Marius and Siegwart, Roland and Nieto, Juan},
  booktitle={IEEE/RSJ International Conference on Intelligent Robots and Systems (IROS)},
  pages={1366--1373},
  year={2017},
  organization={IEEE}
}

@INPROCEEDINGS{schmid2022,
  author={Schmid, Lukas and Delmerico, Jeffrey and Schönberger, Johannes L. and Nieto, Juan and Pollefeys, Marc and Siegwart, Roland and Cadena, Cesar},
  booktitle={2022 International Conference on Robotics and Automation (ICRA)}, 
  title={\href{https://arxiv.org/pdf/2109.10165}{Panoptic {M}ulti-{TSDF}s: {A} Flexible Representation for Online Multi-resolution Volumetric Mapping and Long-term Dynamic Scene Consistency}}, 
  year={2022},
  pages={8018-8024}
}

@inproceedings{pan2022,
  title={\href{https://ieeexplore.ieee.org/document/9981318}{Voxfield: {N}on-projective signed distance fields for online planning and 3{D} reconstruction}},
  author={Pan, Yue and Kompis, Yves and Bartolomei, Luca and Mascaro, Ruben and Stachniss, Cyrill and Chli, Margarita},
  booktitle={IEEE/RSJ International Conference on Intelligent Robots and Systems},
  pages={5331--5338},
  year={2022},
  organization={IEEE}
}

@article{museth2013,
  title={\href{https://dl.acm.org/doi/10.1145/2487228.2487235}{{VDB}: {H}igh-resolution sparse volumes with dynamic topology}},
  author={Museth, Ken},
  journal={ACM Transactions on Graphics},
  volume={32},
  pages={1--22},
  year={2013},
  publisher={ACM New York, NY, USA}
}

@article{vizzo2022,
  AUTHOR = {Vizzo, Ignacio and Guadagnino, Tiziano and Behley, Jens and Stachniss, Cyrill},
  TITLE = {\href{https://www.mdpi.com/1424-8220/22/3/1296}{{VDBF}usion: {F}lexible and Efficient {TSDF} Integration of Range Sensor Data}},
  JOURNAL = {Sensors},
  VOLUME = {22},
  YEAR = {2022},
  NUMBER = {3},
  ARTICLE-NUMBER = {1296},
  PubMedID = {35162040},
  ISSN = {1424-8220}
}

@misc{chen2023,
    title = {\href{https://github.com/fudan-zvg/Semantic-Segment-Anything}{Semantic Segment Anything}},
    author = {Chen, Jiaqi and Yang, Zeyu and Zhang, Li},
    year = {2023}
}

@inproceedings{curless1996,
  title={\href{https://dl.acm.org/doi/10.1145/237170.237269}{A volumetric method for building complex models from range images}},
  author={Curless, Brian and Levoy, Marc},
  booktitle={23rd Annual Conference on Computer Graphics and Interactive Techniques},
  pages={303--312},
  year={1996}
}

@incollection{museth2021,
  title={\href{https://dl.acm.org/doi/10.1145/3450623.3464653}{Nano{VDB}: {A} {GPU}-friendly and portable {VDB} data structure for real-time rendering and simulation}},
  author={Museth, Ken},
  booktitle={ACM SIGGRAPH 2021 Talks},
  pages={1--2},
  year={2021}
}

@INPROCEEDINGS{besselman2021,
  author={Besselmann, Marvin Grosse and Puck, Lennart and Steffen, Lea and Roennau, Arne and Dillmann, R{\"u}diger},
  booktitle={2021 IEEE 17th International Conference on Automation Science and Engineering (CASE)}, 
  title={\href{https://ieeexplore.ieee.org/document/9551430}{{VDB-Mapping}: {A} High Resolution and Real-Time Capable {3D} Mapping Framework for Versatile Mobile Robots}}, 
  year={2021},
  pages={448-454}
}

@InProceedings{besselman2022,
    author="Besselmann, Marvin Grosse
    and R{\"o}nnau, Arne
    and Dillmann, R{\"u}diger",
    title={\href{https://link.springer.com/chapter/10.1007/978-3-031-15226-9_42}{Remote {VDB}-{M}apping: {A} Level-Based Data Reduction Framework for Distributed Mapping}},
    booktitle="Robotics in Natural Settings",
    year="2022",
    publisher="Springer International Publishing",
    pages="448--459",
    isbn="978-3-031-15226-9"
}

@INPROCEEDINGS{hagmanns2022,
  title={\href{https://arxiv.org/pdf/2211.04067}{Efficient Global Occupancy Mapping for Mobile Robots using Open{VDB}}},
  author={Hagmanns, Raphael and Emter, Thomas and Grosse Besselmann, Marvin and Beyerer, J{\"u}rgen},
  booktitle={Proceedings of the IROS Agile Robotics Workshop},
  year={2022}
}

@INPROCEEDINGS{bai2023,
  author={Bai, Yinlong and Miao, Zhiqiang and Wang, Xiangke and Liu, Yong and Wang, Hesheng and Wang, Yaonan},
  booktitle={IEEE/RSJ International Conference on Intelligent Robots and Systems (IROS)}, 
  title={\href{https://ieeexplore.ieee.org/document/10342123}{{VDB}blox: {A}ccurate and Efficient Distance Fields for Path Planning and Mesh Reconstruction}}, 
  year={2023},
  pages={7187-7194}
}

@article{wilson2024,
  title={{LatentBKI}: Open-Dictionary Continuous Mapping in Visual-Language Latent Spaces With Quantifiable Uncertainty},
  author={Wilson, Joey and Xu, Ruihan and Sun, Yile and Ewen, Parker and Zhu, Minghan and Barton, Kira and Ghaffari, Maani},
  journal={IEEE Robotics and Automation Letters},
  year={2025},
  publisher={IEEE}
}

@article{deng2023,
  title={\href{https://ieeexplore.ieee.org/document/10093797}{{SEE}-{CSOM}: {S}harp-edged and efficient continuous semantic occupancy mapping for mobile robots}},
  author={Deng, Yinan and Wang, Meiling and Yang, Yi and Wang, Danwei and Yue, Yufeng},
  journal={IEEE Transactions on Industrial Electronics},
  volume={71},
  number={2},
  pages={1718--1728},
  year={2023},
  publisher={IEEE}
}

@inproceedings{behley2019,
  author = {Jens Behley and Martin Garbade and Andres Milioto and Jan Quenzel and Sven Behnke and Cyriss Stachniss and Juergen Gall},
  title = {\href{https://arxiv.org/pdf/1904.01416}{Semantic{KITTI}: {A} Dataset for Semantic Scene Understanding of {L}i{DAR} Sequences}},
  booktitle = {Proceedings of the IEEE/CVF International Conference on Computer Vision (ICCV)},
  year = {2019}
}

@inproceedings{mccormac2017,
  title={\href{https://openaccess.thecvf.com/content_ICCV_2017/papers/McCormac_SceneNet_RGB-D_Can_ICCV_2017_paper.pdf}{Scenenet {RGB}-{D}: {C}an 5{M} synthetic images beat generic {I}mage{N}et pre-training on indoor segmentation?}},
  author={McCormac, John and Handa, Ankur and Leutenegger, Stefan and Davison, Andrew J},
  booktitle={Proceedings of the IEEE International Conference on Computer Vision (ICCV)},
  pages={2678--2687},
  year={2017}
}

@misc{murphy2007conjugate,
  title={\href{https://www.cs.ubc.ca/~murphyk/Papers/bayesGauss.pdf}{Conjugate {B}ayesian analysis of the {G}aussian distribution}},
  author={Murphy, Kevin P},
  volume={1},
  pages={16},
  year={2007}
}

@article{tu2014,
  title={\href{https://stephentu.github.io/writeups/dirichlet-conjugate-prior.pdf}{The {D}irichlet-{M}ultinomial and {D}irichlet-{C}ategorical models for {B}ayesian inference}},
  author={Tu, Stephen},
  journal={UC Berkeley},
  volume={2},
  year={2014}
}

@INPROCEEDINGS{SemanticFrequent3D,
  author={He, Hu and Upcroft, Ben},
  booktitle={IEEE/RSJ International Conference on Intelligent Robots and Systems (IROS)}, 
  title={\href{https://ieeexplore.ieee.org/document/6696884}{{Nonparametric semantic segmentation for {3D} street scenes}}}, 
  year={2013},
  pages={3697-3703}
}

@ARTICLE{ConvBKI2,
  author={Wilson, Joey and Fu, Yuewei and Friesen, Joshua and Ewen, Parker and Capodieci, Andrew and Jayakumar, Paramsothy and Barton, Kira and Ghaffari, Maani},
  journal={IEEE Transactions on Robotics}, 
  title={\href{https://ieeexplore.ieee.org/document/10663972}{{ConvBKI: Real-Time Probabilistic Semantic Mapping Network With Quantifiable Uncertainty}}}, 
  year={2024},
  volume={40},
  pages={4648-4667}
}

@article{MappingSBKI,
    author = {Gan, Lu and Zhang, Ray and Grizzle, Jessy W. and Eustice, Ryan M. and Ghaffari, Maani},
    journal = {IEEE Robotics and Automation Letters},
    number = {2},
    title = {\href{https://ieeexplore.ieee.org/document/8954837}{Bayesian Spatial Kernel Smoothing for Scalable Dense Semantic Mapping}},
    volume = {5},
    year = {2020}
}

@inproceedings{ewen2024you,
    author = {Ewen, Parker and Chen, Hao and Chen, Yuzhen and Li, Anran and Bagali, Anup and Gunjal, Gitesh and Vasudevan, Ram},
    year = {2024},
    booktitle = {Robotics: Science and Systems},
    title = {\href{https://arxiv.org/pdf/2402.05872}{You’ve Got to Feel It To Believe It: Multi-Modal Bayesian Inference for Semantic and Property Prediction}}
}

@article{walker2022,
  title={\href{https://www.mdpi.com/2072-4292/14/21/5463}{{NanoMap}: {A} {GPU}-accelerated {OpenVDB}-based mapping and simulation package for robotic agents}},
  author={Walker, Violet and Vanegas, Fernando and Gonzalez, Felipe},
  journal={Remote Sensing},
  volume={14},
  number={21},
  pages={5463},
  year={2022},
  publisher={MDPI}
}

@inproceedings{radford2021,
  title={\href{https://proceedings.mlr.press/v139/radford21a/radford21a.pdf}{Learning transferable visual models from natural language supervision}},
  author={Radford, Alec and Kim, Jong Wook and Hallacy, Chris and Ramesh, Aditya and Goh, Gabriel and Agarwal, Sandhini and Sastry, Girish and Askell, Amanda and Mishkin, Pamela and Clark, Jack},
  booktitle={International Conference on Machine Learning},
  pages={8748--8763},
  year={2021},
  organization={PMLR}
}

@article{isaacson2023,
  title={\href{https://ieeexplore.ieee.org/abstract/document/10284988}{{LONER}: Lidar only neural representations for real-time {SLAM}}},
  author={Isaacson, Seth and Kung, Pou-Chun and Ramanagopal, Mani and Vasudevan, Ram and Skinner, Katherine A},
  journal={IEEE Robotics and Automation Letters},
  year={2023},
  publisher={IEEE}
}

@InProceedings{park2022,
    author    = {Park, Hyojin and Yessenbayev, Alan and Singhal, Tushar and Adhikari, Navin Kumar and Zhang, Yizhe and Borse, Shubhankar Mangesh and Cai, Hong and Pandey, Nilesh Prasad and Yin, Fei and Mayer, Frank and Calidas, Balaji and Porikli, Fatih},
    title     = {\href{https://openaccess.thecvf.com/content/CVPR2022/papers/Park_Real-Time_Accurate_and_Consistent_Video_Semantic_Segmentation_via_Unsupervised_Adaptation_CVPR_2022_paper.pdf}{Real-Time, Accurate, and Consistent Video Semantic Segmentation via Unsupervised Adaptation and Cross-Unit Deployment on Mobile Device}},
    booktitle = {Proceedings of the IEEE/CVF Conference on Computer Vision and Pattern Recognition},
    year      = {2022},
    pages     = {21431-21438}
}

@article{williams2024,
  title={\href{https://dl.acm.org/doi/10.1145/3658226}{{fVDB}: A deep-learning framework for sparse, large scale, and high performance spatial intelligence}},
  author={Williams, Francis and Huang, Jiahui and Swartz, Jonathan and Klar, Gergely and Thakkar, Vijay and Cong, Matthew and Ren, Xuanchi and Li, Ruilong and Fuji-Tsang, Clement and Fidler, Sanja and Sifakis, Efychios and Museth, Ken},
  journal={ACM Transactions on Graphics},
  volume={43},
  number={4},
  pages={1--15},
  year={2024},
  publisher={ACM New York, NY, USA}
}

@inproceedings{zhou2021,
  title={\href{https://openaccess.thecvf.com/content/CVPR2021/papers/Zhou_Panoptic-PolarNet_Proposal-Free_LiDAR_Point_Cloud_Panoptic_Segmentation_CVPR_2021_paper.pdf}{Panoptic-Polarnet: Proposal-free lidar point cloud panoptic segmentation}},
  author={Zhou, Zixiang and Zhang, Yang and Foroosh, Hassan},
  booktitle={Proceedings of the IEEE/CVF Conference on Computer Vision and Pattern Recognition (CVPR)},
  pages={13194-13203},
  year={2021}
}

@article{tosi2024,
  title={\href{https://arxiv.org/abs/2402.13255}{How {NeRFs} and {3D} gaussian splatting are reshaping {SLAM}: a survey}},
  author={Tosi, Fabio and Zhang, Youmin and Gong, Ziren and Sandstr{\"o}m, Erik and Mattoccia, Stefano and Oswald, Martin R and Poggi, Matteo},
  journal={arXiv preprint arXiv:2402.13255},
  volume={4},
  pages={1},
  year={2024},
  publisher={Apr}
}

\end{document}